\documentclass[11pt,a4paper]{article}
\usepackage[hyperref]{acl2021}
\usepackage{times}
\usepackage{latexsym}

\usepackage{booktabs}
\usepackage{multirow}
\usepackage{bm}
\usepackage{amsmath}
\usepackage{graphicx}
\usepackage{CJKutf8}
\usepackage{xcolor}
\usepackage{xpinyin}
\usepackage{pifont}

\usepackage{microtype}

\aclfinalcopy 

\newcommand\model{\textsc{ReaLiSe}}

\newcommand{\cjksong}[1]{{{\begin{CJK*}{UTF8}{gbsn}#1\end{CJK*}}}}
\newcommand{\cjkkai}[1]{{{\begin{CJK*}{UTF8}{gkai}#1\end{CJK*}}}}

\title{Read, Listen, and See: Leveraging Multimodal Information \\ Helps Chinese Spell Checking}

\author{Heng-Da Xu$^1$\thanks{\; Heng-Da Xu and Zhongli Li contributed equally. Work is done during internship at Tencent Cloud Xiaowei. Qingyu Zhou is the corresponding author.},~~Zhongli Li$^{2*}$,~~Qingyu Zhou$^2$,~~Chao Li$^2$,~~Zizhen Wang$^2$,\\\textbf{Yunbo Cao}$^2$,~~\textbf{Heyan Huang}$^1$,~~\textbf{Xian-Ling Mao}$^1$
 \\
  $^1$Beijing Institute of Technology \\
  $^2$Tencent Cloud Xiaowei \\
   \texttt{\{xuhengda,hhy63,maoxl\}@bit.edu.cn} \\
   \texttt{\{neutrali,qingyuzhou,diegoli,zizhenwang,yunbocao\}@tencent.com}
  }

\date{}

\begin{document}
\maketitle
\begin{abstract}
Chinese Spell Checking (CSC) aims to detect and correct erroneous characters for user-generated text in Chinese language.
Most of the Chinese spelling errors are misused semantically, phonetically or graphically similar characters.
Previous attempts notice this phenomenon and try to utilize the similarity relationship for this task.
However, these methods use either heuristics or handcrafted confusion sets to predict the correct character.
In this paper, we propose a Chinese spell checker called \model{}, by directly leveraging the multimodal information of the Chinese characters.
The \model{} model tackles the CSC task by (1) capturing the semantic, phonetic and graphic information of the input characters, and (2) selectively mixing the information in these modalities to predict the correct output.
Experiments\footnote{Code and model are available at \url{https://github.com/DaDaMrX/ReaLiSe}.} on the SIGHAN benchmarks show that the proposed model outperforms strong baselines by a large margin.
\end{abstract}

\section{Introduction}

The Chinese Spell Checking (CSC) task aims to identify erroneous characters and generate candidates for correction. It has attracted much research attention, due to its fundamental and wide applications such as search query correction~\cite{app-search,gao-etal-2010}, optical character recognition (OCR)~\cite{afli2016using}, automatic essay scoring~\cite{dong2016automatic}.
Recently, rapid progress~\citep{softmask-spell, spellgcn} has been made on this task, because of the success of large pretrained language models~\citep{bert,roberta,xlnet}.

\begin{CJK*}{UTF8}{gbsn}

\begin{table}[th]
\centering
\small

\begin{tabular}{p{0.55cm}|p{5.85cm}p{0.05cm}}
\toprule

\multicolumn{3}{l}{Phonetically Similar Case} \\ 

\midrule
 
Sent.  & \multicolumn{2}{l}{晚饭后他递给我一\textcolor{red}{\cjkkai{平}(\textit{\pinyin{ping2}, flat)}}红酒。} \\ 

\midrule
& 晚饭后他递给我一\textcolor{orange}{\cjkkai{杯}(\textit{\pinyin{bei1}, cup})}红酒。 & \ding{55} \\
Cand. & 晚饭后他递给我一\textcolor{blue}{\cjkkai{瓶}(\textit{\pinyin{ping2}, bottle})}红酒。 & \ding{51} \\
& 晚饭后他递给我一\textcolor{orange}{\cjkkai{箱}(\textit{\pinyin{xiang1}, box})}红酒。 & \ding{55} \\

\midrule

Trans. & \multicolumn{2}{l}{ He handed me a \textcolor{blue}{\underline{\textit{bottle}}} of red wine after dinner.} \\ 
\midrule \midrule

\multicolumn{3}{l}{Graphically Similar Case} \\ 

\midrule

Sent. & \multicolumn{2}{l}{每天放学我都会\textcolor{red}{\cjkkai{轻}(\textit{\pinyin{qing1}, light})}过这片树林。} \\
 
\midrule

 & 每天放学我都会\textcolor{orange}{\cjkkai{路}(\textit{\pinyin{lu4}, pass})}过这片树林。 & \ding{55} \\
Cand. & 每天放学我都会\textcolor{blue}{\cjkkai{经}(\textit{\pinyin{jing1}, go})}过这片树林。 & \ding{51} \\
 & 每天放学我都会\textcolor{orange}{\cjkkai{走}(\textit{\pinyin{zou3}, walk})}过这片树林。 & \ding{55} \\
 
\midrule 

Trans. & \multicolumn{2}{l}{I \textcolor{blue}{\underline{\textit{go}}} through this wood every day after school.} \\


\bottomrule
\end{tabular}

\caption{Two examples of Chinese spelling errors and their candidate corrections. ``Sent./Cand./Trans." are short for sentence/candidates/translation respectively. The \textcolor{red}{wrong}/\textcolor{orange}{candidate}/\textcolor{blue}{correct} characters with their pronunciation and translation are in \textcolor{red}{red}/\textcolor{orange}{orange}/\textcolor{blue}{blue} color.}
\label{tab:intro}
\end{table}

\end{CJK*}

In alphabetic languages such as English, spelling errors often occur owing to one or more wrong characters, resulting in the written word not in the dictionary problem~\citep{en-sec}.
However, 
Chinese characters are valid if they can be typed in computer systems, which causes that the spelling errors are de facto misused characters in the context of computer-based language processing.
Considering the formation of Chinese characters, a few of them were originally pictograms or phono-semantic compound characters~\citep{norman1988chinese}.
Thus, in Chinese, the spelling errors are not only the misused characters with confusing semantic meaning, but also the characters which are phonetically or graphically similar~\cite{vis-pho-ratio,similar-character}. Table~\ref{tab:intro} shows two examples of Chinese spelling error.
In the first example, phonetic information of ``\cjksong{平}'' (flat) is needed to get the correct character ``\cjksong{瓶}'' (bottle) since they share the same pronunciation ``\pinyin{ping2}''.
The second example needs not only phonetic, but also graphic information of the erroneous character ``\cjksong{轻}'' (light).
The correct one, ``\cjksong{经}'' (go), has the same right radical as ``\cjksong{轻}'' and similar pronunciation (``\pinyin{qing1}'' and ``\pinyin{jing1}'').
Therefore, considering the intrinsic nature of Chinese, it is essential to leverage the phonetic and graphic knowledge of the Chinese characters along with the textual semantics for the CSC task.

In this paper, we propose \textsc{ReaLiSe} (\textbf{Rea}d, \textbf{Li}sten, and \textbf{Se}e), a Chinese spell checker which leverages the semantic, phonetic and graphic information to correct the spelling errors.
The \model{} model employs three encoders to learn informative representations from textual, acoustic and visual modalities.
First, BERT~\citep{bert} is adopted as the backbone of the semantic encoder to capture the textual information. 
For the acoustic modality, \textit{Hanyu Pinyin} (pinyin), the romanization spelling system for the sounds of Chinese characters, is used as the phonetic features. We design a hierarchical encoder to process the pinyin letters at the character-level and the sentence-level.
Meanwhile, for the visual modality, we build character images with multiple channels as the graphic features, where each channel corresponds to a specific Chinese font.
Then, we use ResNet~\citep{resnet} blocks to encode the images to get the graphic representation of characters.

With the representation of three different modalities, one challenge is how to fuse them into one compact multimodal representation.
To this end, a selective modality fusion mechanism is designed to control how much information of each modality can flow to the mixed representation.
Furthermore, as the pretrain-finetune procedure has been proven to be effective on various NLP tasks~\citep{bert,unilm,ernie2}, we propose to pretrain the phonetic and the graphic encoders by predicting the correct character given input in the corresponding modality.

We conduct experiments on the SIGHAN benchmarks~\citep{sighan13, sighan14, sighan15}.
By leveraging multimodal information, \model{} outperforms all previous state-of-the-art models by a large margin.
Compared to previous methods using confusion set~\cite{conf} to capture the character similarity relationships, such as the SOTA SpellGCN~\citep{spellgcn}, \model{} achieves an averaging 2.4\% and 2.6\%  F1 improvements at detection-level and correction-level. 
Further analysis shows that our model performs better on the errors which are not defined in the handcrafted confusion sets.
This indicates that leveraging the phonetic and graphic information of Chinese characters can better capture the easily-misused characters.

To summarize, the contributions of this paper include:
(i) we propose to leverage phonetic and graphic information of Chinese characters besides textual semantics for the CSC task;
(ii) we introduce the selective fusion mechanism to integrate multimodal information;
(iii) we propose acoustic and visual pretraining tasks to further boost the model performance;
(iv) to the best of our knowledge, the proposed \model{} model achieves the best results on the SIGHAN CSC  benchmarks.

\begin{figure*}[htbp]
    \centering
    \includegraphics[width=1.0\textwidth]{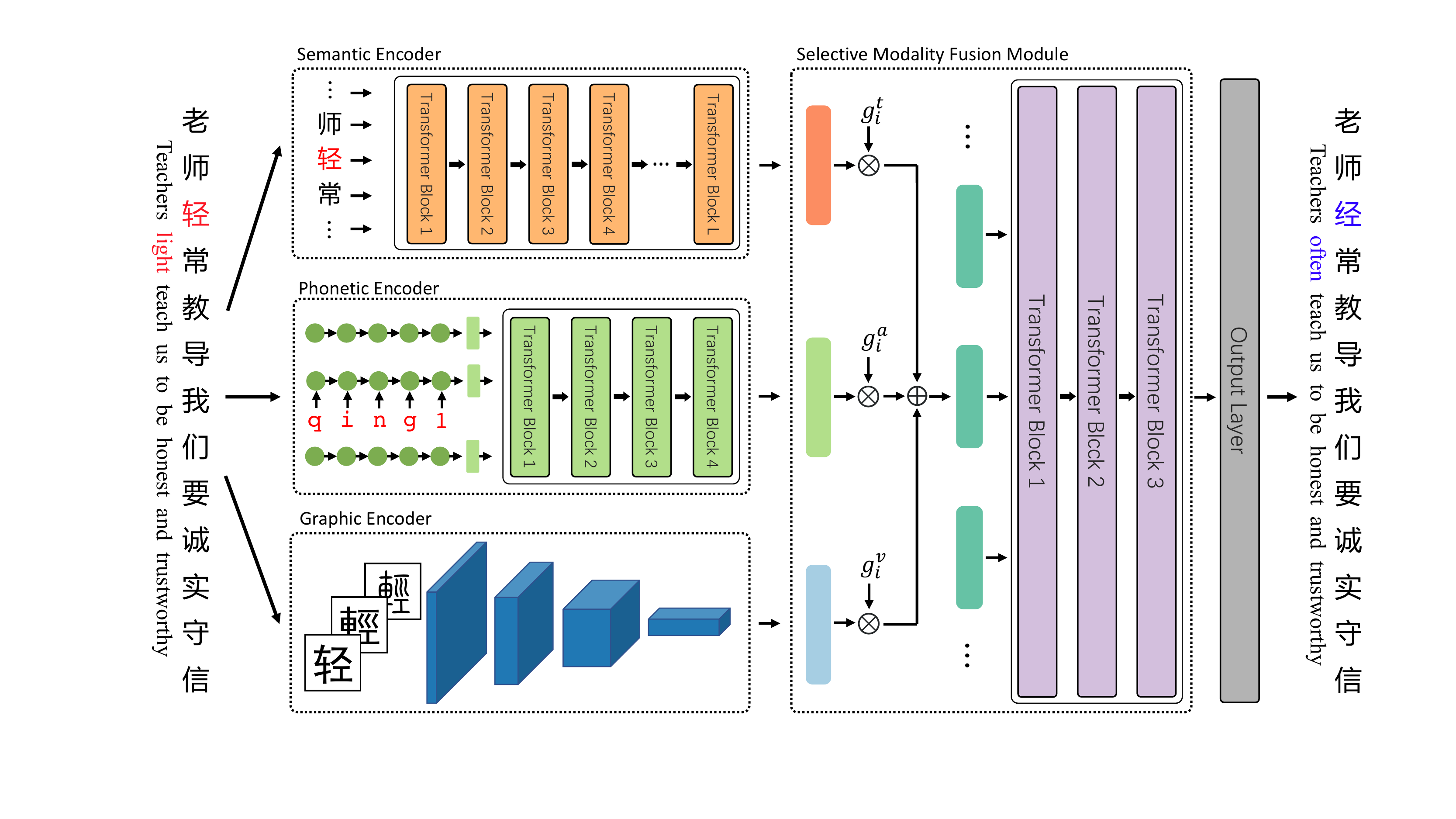}
    \caption{Architecture overview of the \model{} model. 
    The semantic, phonetic and graphic encoders, are used to capture the information in textual, acoustic and visual modalities.
    The fusion module selectively fuses the information from three encoders.
    In the example input, to correct the erroneous character, ``\cjksong{轻}" (\pinyin{qing1}, light), we need not only the contextual text information, but also the phonetic and graphic information of the character itself.
    }
    \label{fig:model}
\end{figure*}

\section{Related Work}

\subsection{Chinese Spell Checking}

The CSC task is to detect and correct spelling errors in Chinese sentences. 
Early works design various rules to deal with different errors~\citep{kuas, ntou}. Next, traditional machine learning algorithms are brought to this field, such as Conditional Random Field and Hidden Markov Model~\citep{nctu, HANSpeller++}. Then, neural-based methods have made great progress in CSC. \citet{spell-corpus} treat the CSC task as a sequence labeling problem, and use a bidirectional LSTM to predict the correct characters. With the great success of large pretrained language models (e.g., BERT~\cite{bert}), \citet{FASPell} propose the FASpell model, which use a BERT-based denoising autoencoder to generate candidate characters and uses some empirical measures to select the most likely ones. Besides, the Soft-Masked BERT model~\citep{softmask-spell} leverages a cascading architecture where GRU is used to detect the erroneous positions and BERT is used to predict correct characters.

Previous works~\cite{yu2014chinese,pointer-spell,spellgcn} using handcrafted Chinese character confusion set~\citep{conf} aim to correct the errors by discovering the similarity of the easily-misused characters.
\citet{pointer-spell} leverage the pointer network~\cite{vinyals2015pointer} by picking the correct character from the confusion set.
\citet{spellgcn} propose a SpellGCN model which models the character similarity through Graph Convolution Network (GCNs)~\citep{gcn} on the confusion set.
However, the character confusion set is predefined and fixed, which cannot cover all the similarity relations, nor can it distinguish the divergence in the similarity of Chinese characters. 
In this work, we discard the predefined confusion set and directly use the multi-modal information to discover the subtle similarity relationship between all Chinese characters.

\subsection{Multimodal Learning}


There has been much research to integrate information from different modalities to achieve better performance. 
Tasks such as Multimodal Sentiment Analysis~\citep{mm-senti, msa-2}, Visual Question Answering~\citep{vqa, vqa-2} and Multimodal Machine Translation~\citep{mmt-1, mmt-2} have made much progress.
Recently, multimodal pretraining models have been proposed, such as VL-BERT~\citep{vl-bert}, Unicoder-VL~\citep{uni-vl}, and LXMERT~\citep{lxmert}. 
In order to incorporate the visual information of Chinese characters into language models, \citet{glyce} design a Tianzige-CNN to facilitate some NLP tasks, such as named entity recognition and sentence classification. To the best of our knowledge, this paper is the first work to leverage multimodal information to tackle the CSC task.

\section{The \model{} Model}

In this section, we introduce the \model{} model, which utilizes the semantic, phonetic, and graphic information to distinguish the similarities of Chinese characters and correct the spelling errors.
As shown in Figure~\ref{fig:model}, multiple encoders are firstly employed to capture valuable information from textual, acoustic and visual modalities. Then, we develop a selective modality fusion module to obtain the context-aware multimodal representations. Finally, the output layer predicts the probabilities of error corrections.

\subsection{The Semantic Encoder}

We adopt BERT~\citep{bert} as the backbone of the semantic encoder. 
BERT provides rich contextual word representation with the unsupervised pretraining on large corpora.

The input tokens $\mathbf{X}=(x_1, \dots, x_N)$ are first projected into $\mathbf{H}^t_0$ through the input embedding.
Then the computation of Transformer~\citep{transformer} encoder layers can be formulated as:
\begin{equation}
    \mathbf{H}^t_l = \text{Transformer}_l (\mathbf{H}^t_{l-1}), l \in [1, L]
\end{equation}
where $L$ is the number of Transformer layers. 
Each layer consists of a multi-head attention module and a feed-forward network with the residual connection~\citep{resnet} and layer normalization~\citep{layer_norm}. 
The output of the last layer $\mathbf{H}^t=\mathbf{H}^t_{L}=(h^t_1, \dots, h^t_N)$ is used as the contextualized semantic representation of the input tokens in textual modality.

\subsection{The Phonetic Encoder}
\textit{Hanyu Pinyin} (pinyin) is the romanization for Chinese to ``spell out'' the sounds of characters.
We use it to calculate the phonetic representation in this paper. The pinyin of a Chinese character consists of three parts: initial, final, and tone. The initial (21 in total) and final (39 in total) are written with letters in the English alphabet. The 5 kinds of tones (take the final ``a'' as an example, \{ \pinyin{a1}, \pinyin{a2}, \pinyin{a3}, \pinyin{a4}, a \}) can be mapped into numbers $\{1,2,3,4,0\}$.
Though the vocabulary size of pinyin for all Chinese characters is a fixed number, we use a sequence of letters in \model{} to capture the subtle phonetic difference between Chinese characters.
For example, the pinyin of ``\cjksong{中}'' (middle) and ``\cjksong{棕}'' (brown) are ``\pinyin{zhong1}'' and ``\pinyin{zong1}'' respectively. The two characters have very similar sounds but quite different meanings. 
We thus represent pinyin as a symbol sequence, e.g., $\{z, h, o, n, g, 1\}$ for ``\cjksong{中}''. 
We denote the pinyin of the $i$-th character in the input sentence as $\bm{p}_i = (p_{i,1}, \dots, p_{i,|\bm{p}_i|})$, where $|\bm{p}_i|$ is the length of pinyin $\bm{p}_i$.

In \model{}, we design a hierarchical phonetic encoder, which consists of a character-level encoder and a sentence-level encoder.

\paragraph{The Character-level Encoder} is to model the basic pronunciation and capture the subtle sound difference between characters. It is a single-layer uni-directional GRU~\citep{gru}, which encodes the pinyin of the $i$-th character $x_i$ as:
\begin{equation}
    \tilde{h}^a_{i,j} = \textrm{GRU}(\tilde{h}^a_{i, j-1}, E(p_{i,j}))
\end{equation}
where $E(p_{i,j})$ is the embedding of the pinyin symbol $p_{i,j}$, and $\tilde{h}^a_{i,j}$ is the $j$-th hidden states of the GRU. The last hidden state is used as the character-level phonetic representation of   $x_i$. 

\paragraph{The Sentence-level Encoder} is a 4-layer Transformer with the same hidden size as the semantic encoder. It is designed to obtain the contextualized phonetic representation for each Chinese character. As the independent phonetic vectors are not distinguished in order, we add the positional embedding to each vector in advance.
Then, we pack these phonetic vectors together, and apply the Transformer layers to calculate the contextualized representation in acoustic modality, denoted as $\mathbf{H}^a = (h^a_1, h^a_2, ..., h^a_N)$. 
Note that owing to the Transformer architecture, this representation is also normalized.

\subsection{The Graphic Encoder}
We apply the ResNet~\citep{resnet} as the graphic encoder.
The graphic encoder has 5 layers of ResNet blocks (denoted as ResNet5) followed by a layer normalization~\cite{layer_norm} operation.
We formulate this procedure as follows:
\begin{equation}
\begin{split}
    \tilde{h}^v_i &= \textrm{ResNet5}(\mathbf{I}_i) \\
    h^v_i &= \textrm{LayerNorm}(\tilde{h}^v_i)
\end{split}
\end{equation}
where $\mathbf{I}_i$ is the image of the $i$-th character $x_i$ in the input sentence, and $\textrm{LayerNorm}$ means layer normalization.

In order to extract graphic information effectively, each block in ResNet5 halves the width and height of the image, and increases the number of channels. Thus, the final output is a vector with the length equal to the number of output channels, i.e., both height and width become $1$.
Furthermore, we set the number of output channels to the hidden size in the semantic encoder for the follow-up modality fusion. 
We denote the representation in visual modality of the input sentence as $\mathbf{H}^v = (h^v_1, h^v_2, \dots, h^v_N)$.

The character image of $x_i$ is read from preset font files.
Since the scripts of Chinese characters have evolved for thousands of years, to capture the graphic relationship between character as much as possible, we select three fonts, namely Gothic typefaces (\cjksong{黑体}, \pinyin{hei1}\pinyin{ti3}) in both Simplified and Traditional Chinese, and Small Seal Script (\cjksong{小篆}, \pinyin{xiao3}\pinyin{zhuan4}).
The three fonts correspond to the three channels of the character images, whose size is set to  $32 \times 32$ pixel. 

\subsection{Selective Modality Fusion Module}

After applying the previously mentioned semantic, phonetic and graphic encoders, we get the representation vectors $\mathbf{H}^t$, $\mathbf{H}^a$ and $\mathbf{H}^v$ in textual, acoustic and visual modalities.
To predict the final correct Chinese characters, we develop a selective modality fusion module to integrate these vectors in different modalities.
This module fuses information in two levels, i.e., character-level and sentence-level.

First, for each modality, a selective gate unit is employed to control how much information can flow to the mixed multimodal representation.
For example, if a character is misspelled due to its similar pronunciation to the correct one, then more information in the acoustic modality should flow into the mixed representation. 
The gate values are computed by a fully-connected layer followed by a sigmoid function.
The inputs include the character representation of three modalities and the mean of the semantic encoder output $\mathbf{H}^t$ to capture the overall semantics of the input sentence. 
Formally, we denote the gate values for the textual, acoustic and visual modalities as $g^t$, $g^a$ and $g^v$. The mixed multimodal representation $\tilde{h}_i$ of the $i$-th character is computed as follows: 
\begin{equation}
\begin{split}
    \Bar{h}^t &= \frac{1}{N} \sum_{i=1}^{N}{h^t_i} \\
    g^t_i &= \sigma(\mathbf{W}^t \cdot [h^t_i, h^a_i, h^v_i, \Bar{h}^t] + b^t) \\
    g^a_i &= \sigma(\mathbf{W}^a \cdot [h^t_i, h^a_i, h^v_i, \Bar{h}^t] + b^a) \\
    g^v_i &= \sigma(\mathbf{W}^v \cdot [h^t_i, h^a_i, h^v_i, \Bar{h}^t] + b^v) \\
    \tilde{h}_i &= g^t_i \cdot h^t_i + g^a_i \cdot h^a_i + g^v_i \cdot h^v_i
\end{split}
\end{equation}
where $\mathbf{W}^t$, $\mathbf{W}^a$, $\mathbf{W}^v$, $b^t$, $b^a$, $b^v$ are learnable parameters, $\sigma$ is the sigmoid function, and $[\cdot]$ means the concatenation of vectors. 

Then, we apply the Transformer to fully learn the semantic, phonetic and visual information at the sentence-level.
The mixed representations of all characters are packed together into $\mathbf{H}_0=[\tilde{h}_1, \tilde{h}_2, ..., \tilde{h}_N]$, and the probability distribution $\hat{y}_i$ of what the $i$-th character should be is derived as:
\begin{equation}
\begin{split}
    \mathbf{H}_l &= \text{Transformer}_l (\mathbf{H}_{l-1}), l \in [1, L'] \\
    \hat{y}_i &= \textrm{softmax}(\mathbf{W}^o h_i + b^o), h_i \in \mathbf{H}_{L'}
\end{split}
\end{equation}
where $L'$ is the number of Transformer layers, $\mathbf{W}^o$ and $b^o$ are learnable parameters. 

\subsection{Acoustic and Visual Pretraining}
While acoustic and visual information is essential to the CSC task, equally important is how to associate them with the correct character.
In order to learn the acoustic-textual and visual-textual relationships, we propose to pretrain the phonetic and the graphic encoders. 

For the phonetic encoder, we design an Input Method pretraining objective, that the encoder should recover the Chinese character sequence given the input pinyin sequence. This is what the Chinese input methods do. 
We add a linear layer on the top of the encoder to transform the hidden states to the probability distributions over the Chinese character vocabulary. 
We pretrain the phonetic encoder with the pinyin of the sentences with spelling errors in the training data, and make it recover the character sequences without spelling errors. 

For the graphic encoder, we design an Optical Character Recognition (OCR) pretraining objective.
Given the Chinese character images, the graphic encoder learns the visual information to predict the corresponding character over the Chinese character vocabulary.
This is like what the OCR task does, but our recognition is only conducted on the character level and typed scripts. During the pretraining, we also add a linear layer on the top to perform the classification. 

Finally, we load the pretrained weights of the semantic encoder, phonetic encoder, and graphic encoder, and conduct the final training process with the CSC training data.


\begin{table}[t]
\small
\centering

\begin{tabular}{@{}lrrr@{}}
\toprule
Training Set & \#Sent & Avg. Length & \#Errors \\ \midrule
SIGHAN13 & 700 & 41.8 & 343 \\
SIGHAN14 & 3,437 & 49.6 & 5,122 \\
SIGHAN15 & 2,338 & 31.3 & 3,037 \\
Wang271K & 271,329 & 42.6 & 381,962 \\ \midrule
Total & 277,804 & 42.6 & 390464 \\ \midrule \midrule
Test Set & \#Sent & Avg. Length & \#Errors \\ \midrule
SIGHAN13 & 1,000 & 74.3 & 1,224 \\
SIGHAN14 & 1,062 & 50.0 & 771 \\
SIGHAN15 & 1,100 & 30.6 & 703 \\ \midrule
Total & 3,162 & 50.9 & 2,698 \\ \bottomrule
\end{tabular}

\caption{Statistics of the used datasets. All the training data are merged to train the \model{} model. The test sets are used separately to evaluate the model performance.}
\label{tab-dataset}
\end{table}

\begin{table*}[th]
\small
\centering

\begin{tabular}{@{}c|l|rrrr|rrrr@{}}
\toprule
\multirow{2}{*}{Dataset} & \multicolumn{1}{c|}{\multirow{2}{*}{Method}} & \multicolumn{4}{c|}{Detection Level} & \multicolumn{4}{c}{Correction Level} \\
 & \multicolumn{1}{c|}{} & Acc & Pre & Rec & F1 & Acc & Pre & Rec & F1 \\ \midrule
 
\multicolumn{1}{l|}{\multirow{6}{*}{SIGHAN13}} & Sequence Labeling~\citep{spell-corpus} & - & 54.0 & 69.3 & 60.7 & - & - & - & 52.1 \\
\multicolumn{1}{l|}{} & FASpell~\citep{FASPell} & 63.1 & 76.2 & 63.2 & 69.1 & 60.5 & 73.1 & 60.5 & 66.2 \\
\multicolumn{1}{l|}{} & BERT~\citep{spellgcn} & - & 79.0 & 72.8 & 75.8 & - & 77.7 & 71.6 & 74.6 \\
\multicolumn{1}{l|}{} & SpellGCN~\citep{spellgcn} & - & 80.1 & 74.4 & 77.2 & - & 78.3 & 72.7 & 75.4 \\ 
\multicolumn{1}{l|}{} & SpellGCN$^{\dagger}$~(Our reimplementation) & 78.8 & 85.7 & 78.8 & 82.1 & 77.8 & 84.6 & 77.8 & 81.0 \\ 
\cmidrule(l){2-10} 
\multicolumn{1}{l|}{} & BERT$^{\dagger}$ & 77.0 & 85.0 & 77.0 & 80.8 & 75.2 & 83.0 & 75.2 & 78.9 \\
\multicolumn{1}{l|}{} & \model{}$^{\dagger}$ & \textbf{82.7} & \textbf{88.6} & \textbf{82.5} & \textbf{85.4} & \textbf{81.4} & \textbf{87.2} & \textbf{81.2} & \textbf{84.1} \\ 

\midrule

\multicolumn{1}{l|}{\multirow{6}{*}{SIGHAN14}} & Sequence Labeling~\citep{spell-corpus} & - & 51.9 & 66.2 & 58.2 & - & - & - & 56.1 \\
\multicolumn{1}{l|}{} & FASpell~\citep{FASPell} & 70.0 & 61.0 & 53.5 & 57.0 & 69.3 & 59.4 & 52.0 & 55.4 \\
\multicolumn{1}{l|}{} & SpellGCN$$~\citep{spellgcn} & - & 65.1 & 69.5 & 67.2 & - & 63.1 & 67.2 & 65.3 \\ \cmidrule(l){2-10} 
\multicolumn{1}{l|}{} & BERT$$ & 75.7 & 64.5 & 68.6 & 66.5 & 74.6 & 62.4 & 66.3 & 64.3 \\
\multicolumn{1}{l|}{} & \model{}$$ & \textbf{78.4} & \textbf{67.8} & \textbf{71.5} & \textbf{69.6} & \textbf{77.7} & \textbf{66.3} & \textbf{70.0} & \textbf{68.1} \\ 

\midrule

\multicolumn{1}{l|}{\multirow{12}{*}{SIGHAN15}} & KUAS~\citep{kuas} & 53.2 & 57.5 & 24.6 & 34.4 & 51.5 & 53.7 & 21.1 & 30.3 \\
 & NTOU~\citep{ntou} & 42.2 & 42.2 & 41.8 & 42.0 & 39.0 & 38.1 & 35.2 & 36.6 \\
 & NCTU-NTUT~\citep{nctu} & 60.1 & 71.7 & 33.6 & 45.7 & 56.4 & 66.3 & 26.1 & 37.5 \\
 & HanSpeller++~\citep{HANSpeller++} & 70.1 & 80.3 & 53.3 & 64.0 & 69.2 & 79.7 & 51.5 & 62.5 \\
 & LMC~\citep{lmc} & 54.6 & 63.8 & 21.5 & 32.1 & 52.3 & 57.9 & 16.7 & 26.0 \\ 
 \cmidrule(l){2-10} 
 & Sequence Labeling~\citep{spell-corpus} & - & 56.6 & 69.4 & 62.3 & - & - & - & 57.1 \\
 & FASpell~\citep{FASPell} & 74.2 & 67.6 & 60.0 & 63.5 & 73.7 & 66.6 & 59.1 & 62.6 \\
 & Soft-Masked BERT~\citep{softmask-spell} & 80.9 & 73.7 & 73.2 & 73.5 & 77.4 & 66.7 & 66.2 & 66.4 \\
 & SpellGCN$$~\citep{spellgcn} & - & 74.8 & 80.7 & 77.7 & - & 72.1 & 77.7 & 75.9 \\ \cmidrule(l){2-10} 
 & BERT$$ & 82.4 & 74.2 & 78.0 & 76.1 & 81.0 & 71.6 & 75.3 & 73.4 \\
 & \model{}$$ & \textbf{84.7} & \textbf{77.3} & \textbf{81.3} & \textbf{79.3} & \textbf{84.0} & \textbf{75.9} & \textbf{79.9} & \textbf{77.8} \\

\bottomrule
\end{tabular}

\caption{The performance of our model and all baseline models on SIGHAN test sets.
The ``$\dagger$" symbol means we apply  post-processing (Section~\ref{ssec:details}) to the model outputs on SIGHAN13. Results of \model{} on all SIGHAN test sets outperforms all the corresponding baselines with a significance level $ p < 0.05$.}
\label{tab:score}
\end{table*}

\section{Experiments}

In this section, we introduce experimental details and results on the SIGHAN benchmarks~\citep{sighan13,sighan14, sighan15}.
We then verify the effectiveness of our model by conducting ablation studies and analyses.

\subsection{Data and Metrics}

Following previous works~\citep{pointer-spell, spellgcn}, we use the SIGHAN training data and the generated pseudo data (\citealp{spell-corpus}, denoted as Wang271K) as the training set. We evaluate our model on the SIGHAN test sets in 2013, 2014 and 2015 (denoted as SIGHAN13, SIGHAN14 and SIGHAN15). Table~\ref{tab-dataset} shows the data statistics. Originally, the SIGHAN datasets are in the Traditional Chinese.
Following previous works~\citep{pointer-spell, spellgcn,softmask-spell}, we convert them to the Simplified Chinese using the OpenCC  tool\footnote{\url{https://github.com/BYVoid/OpenCC}}. 


Results are reported at the detection level and the correction level.
At the detection level, a sentence is considered to be correct if and only if all the spelling errors in the sentence are detected successfully. 
At the correction level, the model must not only detect but also correct all the erroneous characters to the right ones.
We report the accuracy, precision, recall and F1 scores on both levels.

\subsection{Implementation Details}
\label{ssec:details}
The \model{} model is implemented using PyTorch framework~\citep{pytorch} with the Transformer library~\citep{hug-trans}. The architecture of the semantic encoder is same as the BERT\textsubscript{BASE}~\citep{bert} model (i.e. 12 transformer layers with 12 attention heads, hidden size of 768). We initialize the semantic encoder with the weights of BERT-wwm model~\citep{bert-wwm}.
For the phonetic sentence-level encoder, we set the number of layers to 4, and initialize its position embedding with BERT's position embedding.
The selective modality fusion module has 3 transformer layers, i.e., $L'=3$, and the prediction matrix $\mathbf{W}^o$ is tied with the word embedding matrix of the semantic encoder.
All the embeddings and hidden states have the dimension of 768. 
We use the Pillow library to extract the Chinese character images. When processing the special tokens (e.g. \texttt{[CLS]} and \texttt{[SEP]} of BERT), we use the tensor with all zero values as their image inputs.
We train our \model{} model with the AdamW~\citep{adamw} optimizer for 10 epochs. The learning rate is set to 5e-5, the batch size is set to 32, and the model is trained with learning rate warming up and linear decay.

\begin{CJK*}{UTF8}{gbsn}
In the SIGHAN13 test set, the  annotation quality is relatively poor, that quite a lot of the mixed usage of auxiliary ``的'', ``地'', and ``得'' are not annotated~\citep{spellgcn}. Therefore, a well-performed model may obtain bad scores on it. To alleviate the problem, \citet{spellgcn} proposes to continue finetuning the model on the SIGHAN13 training set before testing. We argue that it's not a good practice because it reduces the model performance. Instead, we use a simple and effective post-processing method. We simply remove all the detected and corrected ``的'', ``地'', and ``得'' characters from the model output and then evaluate with the ground truth of SIGHAN13 test set. 

\end{CJK*}



\subsection{Baselines}

We compare \model{} with the following baselines: \textbf{KUAS}~\citep{kuas}, \textbf{NTOU}~\citep{ntou}, \textbf{NCTU-NTUT}~\citep{nctu}, \textbf{HanSpeller++}~\citep{HANSpeller++},  \textbf{LMC}~\citep{lmc} mainly utilize heuristics or traditional machine learning algorithms, such as n-gram language model, Conditional Random Field and Hidden Markov Model. \textbf{Sequence Labeling}~\citep{spell-corpus} treats CSC as a sequence labeling problem and applies a BiLSTM model. \textbf{FASpell}~\citep{FASPell} utilizes a denoising autoencoder (DAE) to generate candidate characters.
\textbf{Soft-Masked BERT}~\citep{softmask-spell} utilizes the detection model to help the correction model learn the right context.
\textbf{SpellGCN}~\citep{spellgcn} incorporates the predefined character confusion sets to the BERT-based correction model through Graph Convolutional Networks (GCNs). 
\textbf{BERT}~\citep{bert} is to directly fine-tune the BERT\textsubscript{BASE} model with the CSC training data.

\subsection{Main Results}
\label{ssec:main-results}
Table~\ref{tab:score} shows the evaluation scores at detection and correction levels on the SIGHAN 13/14/15 test sets.
The \model{} model performs significantly better than all the previous state-of-the-art models on all test sets. It can be seen that, by capturing valuable information from acoustic and visual modalities, \model{} yields consistent gain with a large margin against BERT.
Specifically, at the correction-level, \model{} exceeds BERT by 5.2\% F1 on SIGHAN13, 3.8\% F1 on SIGHAN14, and 4.4\% F1 on SIGHAN15. 
The results on SIGHAN13 are improved significantly with simple post-processing described in Section~\ref{ssec:details}.

There are several successful applications of BERT on the CSC task, such as FASpell and SpellGCN, which also consider the Chinese character similarity. They attempt to calculate the similarity as the confidence of filtering candidates, or construct similarity graphs from predefined confusion sets. 
Instead, in our method, multiple encoders are directly applied to derive more informative representation from the acoustic and visual modalities.
Compared with SpellGCN~\citep{spellgcn}, the SOTA CSC model, our \model{} model achieves an averaging 2.4\% F1 improvements at detection-level and an averaging 2.6\% F1 improvements at correction-level.
This indicates that, compared with other extensions of BERT, the explicit utilization of multimodal information of Chinese characters is more beneficial to the CSC task.




\begin{CJK*}{UTF8}{gbsn}

With the simple post-processing as described in Section~\ref{ssec:details}, results of each model on the SIGHAN13 test set are improved significantly. Compared with BERT and SpellGCN, we can see that, after the post-processing, the \model{} model is ahead of all the baseline models.

\end{CJK*}

\begin{table}[t]
\small
\centering

\begin{tabular}{@{}l|rrrr@{}}
\toprule
\multicolumn{1}{c|}{Model} & Acc & Pre & Rec & F1 \\
\midrule
& \multicolumn{4}{c}{Detection Level} \\
\midrule
~~BERT & 78.4 & 74.6 & 74.5 & 74.5 \\
~~\model{} & 82.0 & 77.9 & 78.5 & 78.1 \\
\quad - Phonetic & 81.2 & 76.4 & 77.7 & 77.0 \\
\quad - Graphic & 81.4 & 77.3 & 77.2 & 77.2 \\
\quad - Multi-Fonts & 81.2 & 76.3 & 77.9 & 77.0 \\
\quad - Pretraining & 81.5 & 76.5 & 78.1 & 77.2 \\
\quad - Selective-Fusion & 81.3 & 76.8 & 77.4 & 77.1 \\ 
\midrule
& \multicolumn{4}{c}{Correction Level} \\
\midrule
~~BERT & 76.9 & 72.3 & 72.3 & 72.3 \\
~~\model{} & 81.0 & 76.5 & 77.0 & 76.7 \\
\quad - Phonetic & 80.2 & 74.8 & 76.1 & 75.4 \\
\quad - Graphic & 80.5 & 75.8 & 75.6 & 75.7 \\
\quad - Multi-Fonts & 80.3 & 74.9 & 76.4 & 75.5 \\
\quad - Pretraining & 80.6 & 75.2 & 76.8 & 75.9 \\
\quad - Selective-Fusion & 80.5 & 75.4 & 76.0 & 75.7 \\ 
\bottomrule
\end{tabular}

\caption{Ablation results of the \model{} model averaged on SIGHAN test sets. We apply the following changes to \model{}: removing the phonetic encoder (- Phonetic), removing the graphic encoder (- Graphic),  using only one font to build the graphic inputs (- Multi-Fonts), removing acoustic and visual pretraining (- Pretraining), replacing the selective modality fusion mechanism with simple summation (- Selective-Fusion).}
\label{tab:ablation}
\end{table}

\subsection{Ablation Study}

We explore the contribution of each component in \model{} by conducting ablation studies with the following settings: 1) removing the phonetic encoder, 2) removing the graphic encoder, 3) using only one font (Gothic typefaces in Simplified Chinese) for the graphic encoder, 4) removing the acoustic and visual pretraining objectives, 5) replacing the selective modality fusion mechanism with simple summation. 

Table~\ref{tab:ablation} shows the averaged scores\footnote{Full ablation results can be found in the Appendix~\ref{apd:ablation}.} on three SIGHAN test sets. 
The main motivation of this paper is to discover the character similarity relationships by incorporating the acoustic and visual information. If removing the phonetic or graphic encoder, we can see that the model performance drops at two levels but still outperforms BERT significantly. This suggests that the checking model can benefit from the multimodal information.
No matter which component we remove, the performance of \model{} drops, which fully demonstrates the effectiveness of each part in our model. 

\begin{figure}[t]
    \centering
    \includegraphics[width=0.465\textwidth]{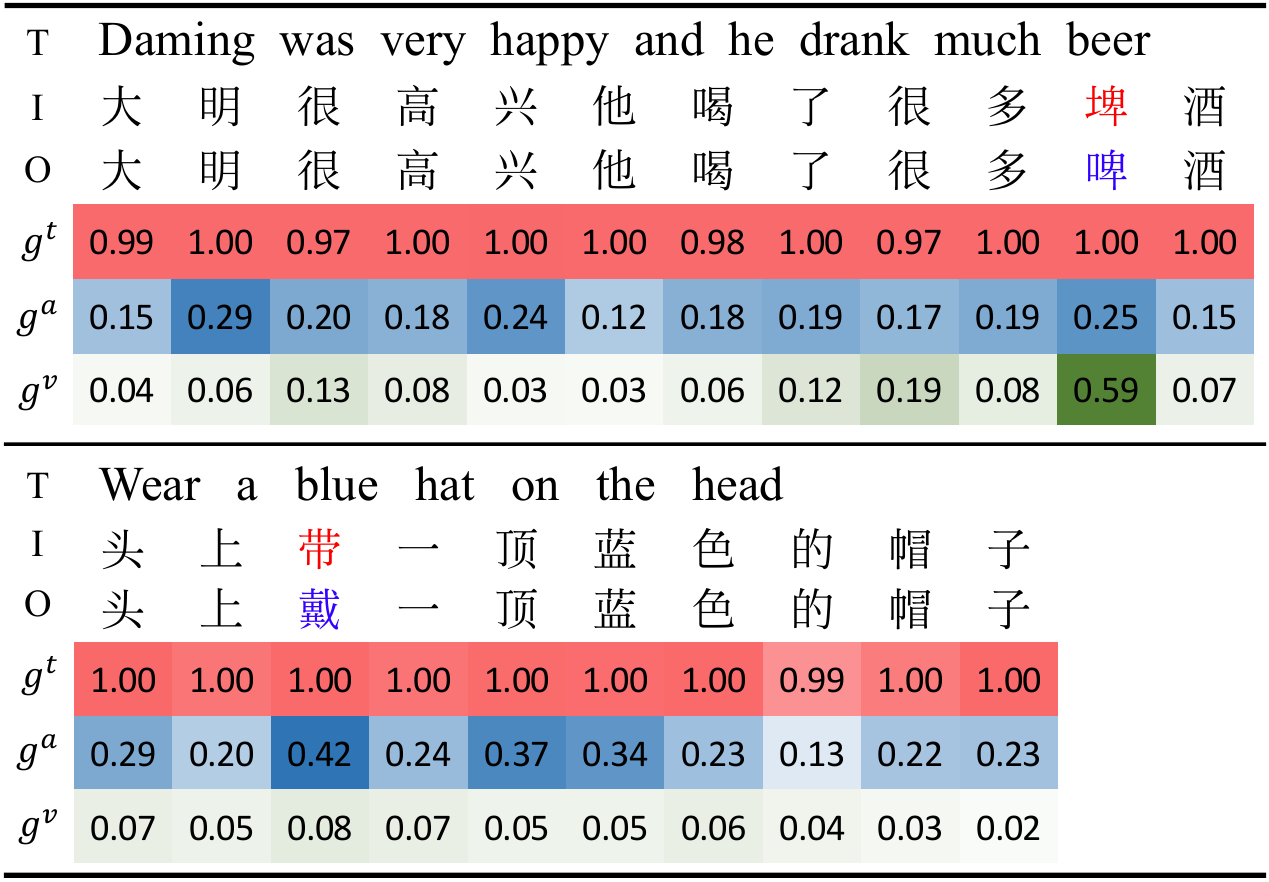}
    \caption{Selective modality fusion visualization. ``I'' is the input sentence. ``O'' is the output of \model{} (also the ground truth), and ``T'' is the translation. $g^t$, $g^a$, $g^v$ are the gate values for the textual, acoustic, and visual modality respectively. We highlight the \textcolor{red}{wrong}/\textcolor{blue}{correct} characters in \textcolor{red}{red}/\textcolor{blue}{blue} color.}
    \label{fig:gate-vis}
\end{figure}

\subsection{Analysis of the Selective Modality Fusion Module}

\begin{CJK*}{UTF8}{gbsn}

Figure~\ref{fig:gate-vis} gives two examples to analyze the selective modality fusion module. 
In the first example, the acoustic and visual selective gate values of ``埤'', i.e. $g^a$ and $g^v$, are much larger than most other characters since ``埤(\pinyin{pi2})'' and ``啤(\pinyin{pi2})'' have the same pronunciation and right radical ``卑''.
This shows that the selective fusion module can judge whether to introduce phonetic or graphic information into the mixed representation.
The second example shows a similar trend for the pronunciation of ``带(\pinyin{dai4})'' and ``戴(\pinyin{dai4})''.
More selective fusion visualization can be found in the Appendix~\ref{apd:fusion}.

Besides, we calculate the averaged gate values of erroneous characters for each modality on SIGHAN15.
The largest one is the textual modality that the value is almost equal to 1.0.
The second one is the acoustic modality that the averaged value is 0.334, and the smallest one is the visual modality that the value is 0.229.
It means that the information from the semantic encoder is the most important for correcting the spelling errors. The acoustic modality is more important than the visual modality, which is consistent with the fact that the spelling errors caused by similar pronunciations are more frequent than errors caused by similar character shapes~\citep{vis-pho-ratio}.

\end{CJK*}

\begin{CJK*}{UTF8}{gbsn}

\begin{table}[t]
\small
\centering

\begin{tabular}{p{0.3cm}p{6.3cm}}
\toprule

\textbf{In:} & 我打算去法国\textcolor{red}{流}行，你要不要跟我一起去？ \\
& I am going to \textcolor{red}{popular} to France, would you like to go with me? \\ 
\midrule
\textbf{Out:} & 我打算去法国\textcolor{blue}{旅}行，你要不要跟我一起去？ \\
& I am going to \textcolor{blue}{travel} to France, would you like to go with me? \\ \midrule \midrule
  
\textbf{In:} & 回国之后，我\textcolor{red}{跟}快去你家。 \\
& After returning home, I will go to your house \textcolor{red}{with}. \\ \midrule
\textbf{Out:} & 回国之后，我\textcolor{blue}{很}快去你家。 \\
& After returning home, I will go to your house \textcolor{blue}{soon}. \\ \bottomrule


\end{tabular}

\caption{Examples of the input and output of our \model{} model. We highlight the \textcolor{red}{wrong}/\textcolor{blue}{correct} characters in \textcolor{red}{red}/\textcolor{blue}{blue} color. 
}
\label{tab:case}
\end{table}

\end{CJK*}

\subsection{Case Study}

\begin{CJK*}{UTF8}{gbsn}

In the first example in Table~\ref{tab:case}, ``流" is the erroneous character. If ignoring the Chinese character similarities, we can find that there are multiple candidate corrections to replace the ``流" character. For instance, we can replace it with ``游" and the English translation is ``\textit{I am going to parade in France}". However, the \model{}'s output is the best correction, because ``流(\pinyin{liu2})'' and ``旅(\pinyin{lv3})'' have a similar pronunciation. In the second example, not only the phonetic information, but also the visual information is important for correcting ``跟(\pinyin{gen1})'' to ``很(\pinyin{hen3})''.
In detail, the two characters share the same final ``en'' in pronunciation, and have the same right radical ``艮''. 

The errors in the above examples are not corrected by SpellGCN, since they are not defined as confusing character pairs in the handcrafted confusion sets~\cite{conf}.
Specifically, in the SIGHAN15 test set, there are 16\% erroneous-corrected character pairs not in the predefined confusion sets. SpellGCN corrects 64.6\% of them but \model{} performs better with 73.5\% correction. Besides, for the easily-confused pairs in the predefined sets, SpellGCN corrects 82.5\% of them and \model{} corrects 85.8\%.
This indicates that leveraging multimodal information of Chinese characters helps the model generalize better in capturing the character similarity relationships.

\end{CJK*}

\section{Conclusion}
In this paper, we propose a model called \model{} for Chinese spell checking.
Since the spelling errors in Chinese are often semantically, phonetically or graphically similar to the correct characters, \model{} leverages information in textual, acoustic and visual modalities to detect and correct the errors.
The \model{} model captures information in these modalities using tailored semantic, phonetic and graphic encoders.
Besides, a selective modality fusion mechanism is proposed to control the information flow of these modalities.
Experiments on the SIGHAN benchmarks show that the proposed \model{} outperforms the baseline models using only textual information by a large margin, which verifies that leveraging acoustic and visual information helps the Chinese spell checking task.

\bibliographystyle{acl_natbib}
\bibliography{main}

\clearpage

\appendix
\section{Appendix}

\begin{figure}[t]
    \centering
    \includegraphics[width=0.48\textwidth]{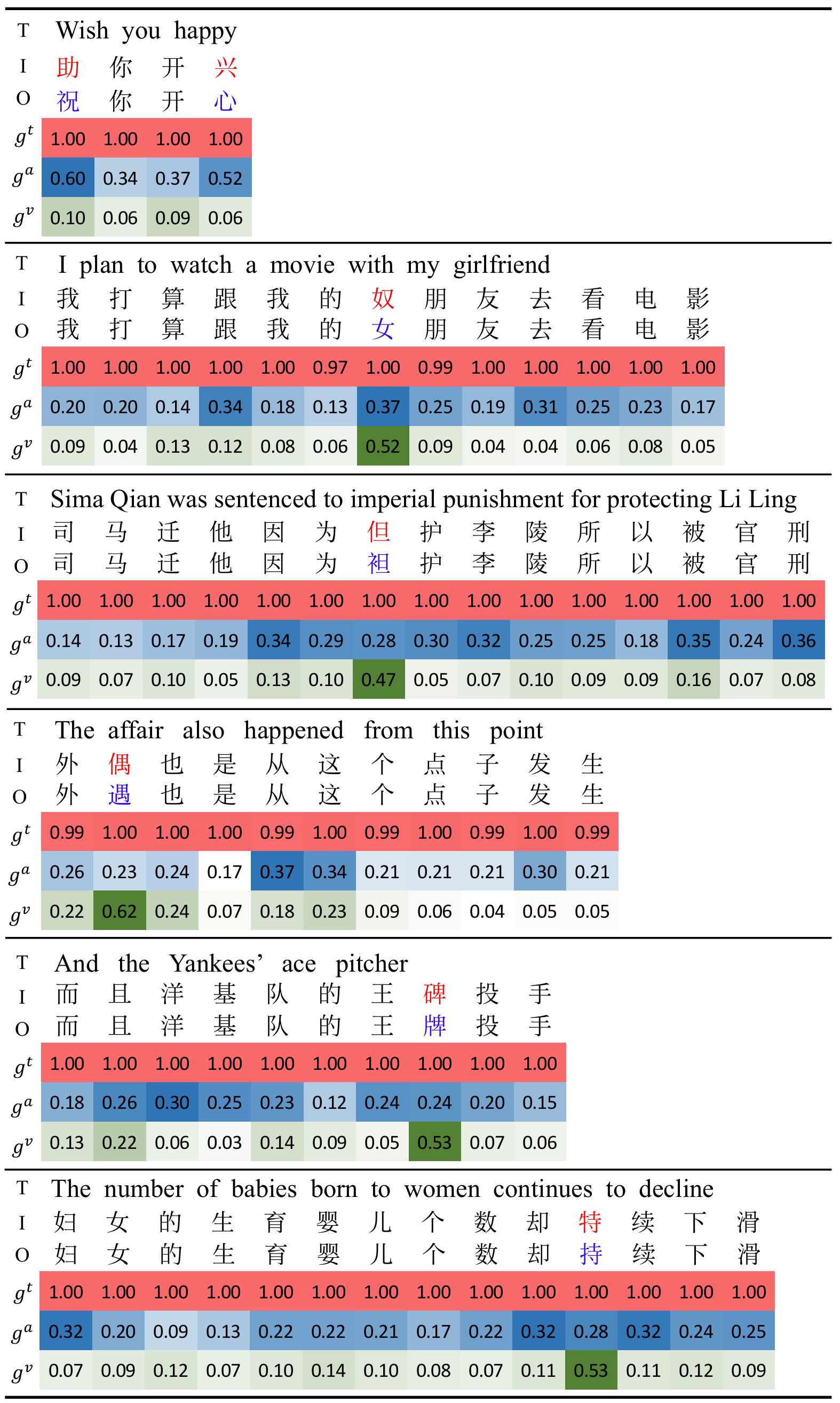}
    \caption{Selective modality fusion visualization. ``I'' is the input sentence. ``O'' is the output of \model{} (also the ground truth), and ``T'' is the translation. $g^t$, $g^a$, $g^v$ are the gate values for the textual, acoustic, and visual modality respectively. We highlight the \textcolor{red}{wrong}/\textcolor{blue}{correct} characters in \textcolor{red}{red}/\textcolor{blue}{blue} color.}
    \label{fig:gate-apd}
\end{figure}

\subsection{Ablation}
\label{apd:ablation}
We conduct an ablation study to verify the effectiveness of the proposed method.
Table~\ref{tab:full-ablation} (on Page~\pageref{tab:full-ablation} in Appendix) shows the detailed ablation results on each SIGHAN test set, where the following settings are conducted:
\begin{enumerate}
    \item \textbf{- Phonetic}: removing the phonetic encoder.
    \item \textbf{- Graphic}: removing the graphic encoder.
    \item \textbf{- Multi-Fonts}: using only one font (Gothic typefaces in Simplified Chinese) for the graphic encoder.
    \item \textbf{- Pretraining}: removing the acoustic and visual pretraining objectives.
    \item \textbf{- Selective-Fusion}: replacing the selective modality fusion mechanism with simple summation.
\end{enumerate}

We can see that, when we remove anything from our model, the \model{} performance drops consistently, and it drops most apparently in the SIGHAN14 test set.
These results suggest that each part of our model is an effective means for boosting the checking performance.

\subsection{Selective Modality Fusion Visualization}
\label{apd:fusion}
We show more examples in Figure~\ref{fig:gate-apd}. We can see that, if the misused characters are phonetically similar to the correct ones, the acoustic gate values tend to be larger, and if they are graphically similar, the visual gate values are larger.

\begin{table*}[t]
\centering

\begin{tabular}{@{}l|@{}l|cccc|cccc}
\toprule
\multirow{2}{*}{~~Dataset} & \multirow{2}{*}{~~Method} & \multicolumn{4}{c|}{Detection Level} & \multicolumn{4}{c}{Correction Level} \\
 & & Acc & Pre & Rec & F1 & Acc & Pre & Rec & F1 \\
\midrule

\multirow{7}{*}{~~SIGHAN13} & ~~BERT & 77.0 & 85.0 & 77.0 & 80.8 & 75.2 & 83.0 & 75.2 & 78.9 \\
& ~~\model{} & 82.7 & 88.6 & 82.5 & 85.4 & 81.4 & 87.2 & 81.2 & 84.1 \\
& \quad- Phonetic & 82.4 & 87.4 & 82.3 & 84.8 & 81.2 & 86.1 & 81.1 & 83.5 \\
& \quad- Graphic  & 82.1 & 88.1 & 82.1 & 85.0 & 80.9 & 86.7 & 80.8 & 83.7 \\
& \quad- Multi-Fonts & 82.2 & 87.5 & 82.2 & 84.8 & 81.2 & 86.4 & 81.2 & 83.7 \\
& \quad- Pretraining & 82.8 & 88.2 & 82.7 & 85.4 & 81.4 & 86.7 & 81.3 & 83.9 \\
& \quad- Selective-Fusion & 82.0 & 87.3 & 82.0 & 84.6 & 81.0 & 86.2 & 81.0 & 83.5  \\

\midrule

\multirow{7}{*}{~~SIGHAN14} & ~~BERT & 75.7 & 64.5 & 68.6 & 66.5 & 74.6 & 62.4 & 66.3 & 64.3 \\
& ~~\model{} & 78.4 & 67.8 & 71.5 & 69.6 & 77.7 & 66.3 & 70.0 & 68.1 \\
& \quad- Phonetic & 77.1 & 65.5 & 69.2 & 67.3 & 76.3 & 63.8 & 67.5 & 65.6 \\
& \quad- Graphic & 78.0 & 67.3 & 69.6 & 68.4 & 77.1 & 65.6 & 67.9 & 66.7 \\
& \quad- Multi-Fonts & 76.9 & 65.0 & 69.6 & 67.2 & 76.2 & 63.6 & 68.1 & 65.7 \\
& \quad- Pretraining & 77.5 & 65.6 & 70.4 & 67.9 & 76.7 & 64.0 & 68.7 & 66.2 \\
& \quad- Selective-Fusion & 77.6 & 66.5 & 69.0 & 67.7 & 76.9 & 64.8 & 67.3 & 66.0 \\

\midrule

\multirow{7}{*}{~~SIGHAN15} & ~~BERT & 82.4 & 74.2 & 78.0 & 76.1 & 81.0 & 71.6 & 75.3 & 73.4 \\
& ~~\model{} & 84.7 & 77.3 & 81.3 & 79.3 & 84.0 & 75.9 & 79.9 & 77.8 \\
& \quad- Phonetic & 84.2 & 76.2 & 81.7 & 78.9 & 83.3 & 74.5 & 79.9 & 77.1 \\
& \quad- Graphic & 84.3 & 76.6 & 79.9 & 78.2 & 83.5 & 75.0 & 78.2 & 76.6 \\
& \quad- Multi-Fonts & 84.5 & 76.5 & 81.9 & 79.1 & 83.5 & 74.6 & 79.9 & 77.1 \\
& \quad- Pretraining & 84.2 & 75.7 & 81.3 & 78.4 & 83.7 & 74.9 & 80.4 & 77.5 \\
& \quad- Selective-Fusion & 84.4 & 76.8 & 81.2 & 78.9 & 83.6 & 75.4 & 79.7 & 77.5 \\
\bottomrule
\end{tabular}
\caption{Ablation results of the \model{} model on each SIGHAN dataset.}
\label{tab:full-ablation}
\end{table*}

\end{document}